\documentclass[twoside]{article}

\usepackage[accepted]{aistats2020}
\usepackage{makecell}
\usepackage{multicol}
\usepackage[T1]{fontenc}
\usepackage{boldline,multirow}
\usepackage{hyperref}       
\usepackage[round]{natbib}
\usepackage{amsfonts}       
\usepackage{nicefrac}       
\usepackage{amsmath}
\usepackage{microtype}      
\usepackage{amssymb}
\usepackage{stmaryrd}
\usepackage{graphicx}
\usepackage{subcaption}
\usepackage{wrapfig}
\usepackage{xcolor}

\newcommand{\rpm}{\raisebox{.2ex}{$\scriptstyle\pm$}}

\DeclareMathOperator*{\argminA}{arg\,min}
\DeclareMathOperator*{\argmaxA}{arg\,max}
\begin{document}
%

%

\twocolumn[

\aistatstitle{Regularization via Structural Label Smoothing}

\aistatsauthor{ Weizhi Li \And Gautam Dasarathy \And  Visar Berisha }

\aistatsaddress{ Arizona State University \And  Arizona State University \And Arizona State University } ]

\begin{abstract}
Regularization is an effective way to promote the generalization performance of machine learning models. In this paper, we focus on label smoothing, a form of output distribution regularization that prevents overfitting of a neural network by softening the ground-truth labels in the training data in an attempt to penalize overconfident outputs. Existing approaches typically use cross-validation to impose this smoothing, which is uniform across all training data. In this paper, we show that such label smoothing imposes a quantifiable bias in the Bayes error rate of the training data, with regions of the feature space with high overlap and low marginal likelihood having a lower bias and regions of low overlap and high marginal likelihood having a higher bias. These theoretical results motivate a simple objective function for data-dependent smoothing to mitigate the potential negative consequences of the operation while maintaining its desirable properties as a regularizer. We call this approach Structural Label Smoothing (SLS). We implement SLS and empirically validate on several synthetic and benchmark datasets (including the CIFAR-100). The results confirm our theoretical insights and demonstrate the effectiveness of the proposed method in comparison to traditional label smoothing.
\end{abstract}
\vspace{-0.2cm}
\section{Introduction}
Regularization methods such as dropout~\citep{srivastava2014dropout}, weight decay~\citep{krogh1992simple}, $\ell_1$-penalization~\citep{tibshirani1996regression, yuan2006model,scardapane2017group}, and batch normalization~\citep{ioffe2015batch} have proven to be an effective means of improving the generalization ability of neural networks by constraining the weights of the model or acting on hidden activations. However, these methods are intricately tied to the specific parameterization and/or the loss criterion of the model in question. A set of complementary approaches penalizes the output distribution of the neural network by either implicitly or explicitly modifying the training labels. Examples of these methods include label smoothing~\citep{szegedy2016rethinking}, label flipping~\citep{xie2016disturblabel}, and low-entropy output distribution penalization~\citep{pereyra2017regularizing}. The implementation of this family of methods has the advantage of being invariant to the parameterization of the network and to the loss criterion used to identify the optimal weights. As such, these regularizers can be implemented as wrappers around any existing optimizer without requiring modifications to the loss function.  
\begin{figure}[t]
\centering
\begin{subfigure}{.23\linewidth}
  \centering
  \includegraphics[width=2cm]{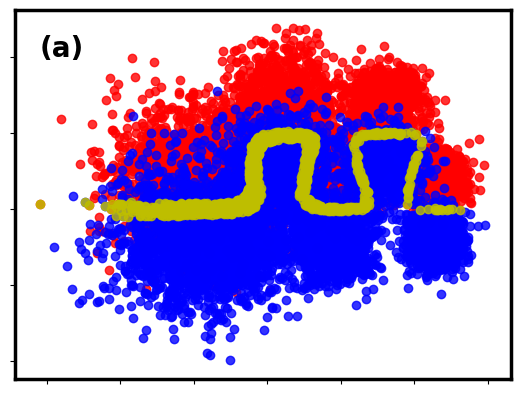}\\
\end{subfigure}
\begin{subfigure}{.23\linewidth}
  \centering
  \includegraphics[width=2cm]{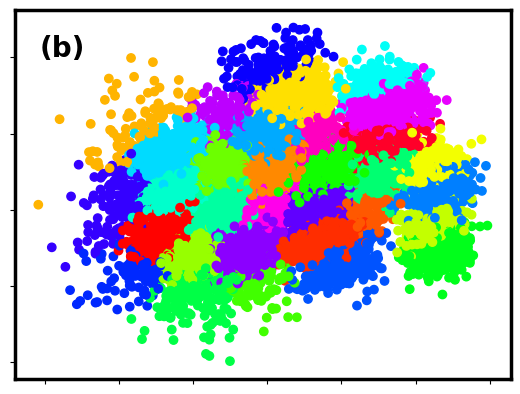}\\
\end{subfigure}
\begin{subfigure}{.23\linewidth}
  \centering
  \includegraphics[width=2cm]{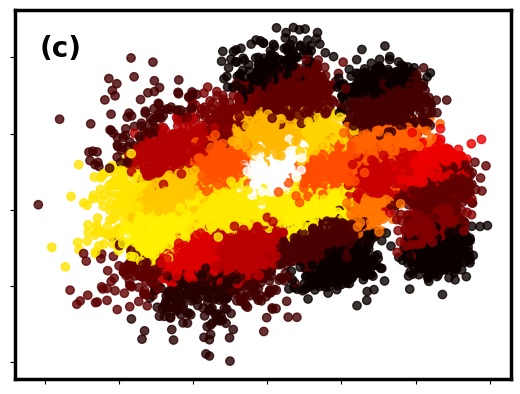}\\
\end{subfigure}
\begin{subfigure}{.27\linewidth}
  \centering
  \includegraphics[width=2.25cm, height=1.51cm]{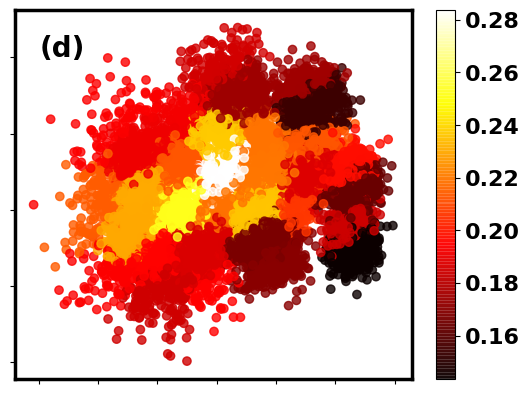}\\
\end{subfigure}
\caption{\footnotesize{Illustration of the smoothing strength inference: (a) Original training data with optimal decision boundary in yellow $\shortrightarrow$ (b) Cluster the training data $\shortrightarrow$ (c) Estimate the BER of each cluster $\shortrightarrow$ (d) Infer the smoothing strength imposed on each cluster. Clusters near the optimal decision boundary are regularized with stronger smoothing strengths.}}
\vspace{-0.7cm}
\label{Fig1:highlevel}
\end{figure}
The existing literature on output distribution regularization ~\citep{szegedy2016rethinking, xie2016disturblabel, pereyra2017regularizing} does not consider the effect of label modifications on the original data distribution. In this paper, we focus on label smoothing since label flipping and entropy penalization can be cast as label smoothing methods. In traditional label smoothing, the level of smoothing is chosen to be uniform across all training samples and is typically chosen via cross-validation. This runs the risk of regularizing certain parts of the feature space too aggressively while not sufficiently regularizing others; e.g., training samples far from the decision boundary may not require the same level of output distribution regularization as those that are near the boundary. This also runs the risk of softening the labels too aggressively and biasing the learning algorithm; in fact, a line of work~\citep{natarajan2013learning, frenay2013classification, li2017noise, xiao2015learning} experimentally shows the adverse impact of label noise on model performance.

In this paper, we first analyze the effects of label smoothing on the training data distribution and then derive an algorithm for optimal, data-dependent smoothing dubbed \emph{structural label smoothing} (SLS). We first derive a loose constraint on how aggressively we can smooth labels before we change the optimal decision boundary between classes. Then we derive an expression for the Bayes error rate (BER) bias in the training data induced by label smoothing. Finally, we use these results to devise a simple objective function to estimate a distribution of smoothing strength values over different regions of the feature space that mitigates this bias. A high-level overview of the process for deriving the smoothing strength distribution over different regions in the feature space is illustrated in Figure~\ref{Fig1:highlevel}.

There are several benefits to this approach: ($1$) The smoothing strength is computed before network training thus SLS circumvents additional computation during training. ($2$) The smoothing strength is correlated with the amount of overlap between class distributions and thus is consistent with the intuition that regions of greater class overlap require higher regularization strength. ($3$) It can be implemented as a wrapper around any existing optimizer as it does not require a modification of the cost function, i.e. only the training label values are modified. We evaluate the SLS against the uniform label smoothing (ULS) from~\citep{szegedy2016rethinking} on synthetic data, Higgs, SVHN, CIFAR-10 and CIFAR-100 datasets. The results show the effectiveness of SLS when compared against traditional ULS.
\vspace{-0.2cm}
\section{Related Work}
\label{sec:relatedwork}
\textbf{Related work on regularization} The previous work most relevant to our approach is the existing work on output distribution regularization that we describe in the introduction~\citep{szegedy2016rethinking, xie2016disturblabel, pereyra2017regularizing}. Other relevant work includes the paper in~\citep{xue2011structural} that directly exploits the structure of the feature space and introduces the concept of ``structural granularity'', with which they propose a structural-regularized support vector machine (SR-SVM) for classification problems. However, their work is focused on SVMs and does not readily extend to other learning algorithms. Furthermore, in contrast to our proposed approach, their approach requires a modification to the underlying cost function. 

Several other approaches to regularization use the structure of the data either implicitly or explicitly. A family of models regularizes implicitly by using the complexity of the class decision boundary. Examples include pairing logits of training samples with either adversarial samples or other clean
samples~\citep{kannan2018adversarial}, constraining the curvature of decision boundary for a classifier~\citep{moosavi2018robustness}, or relating dropout to Rademacher complexity~\citep{zhai2018adaptive}. Another family of regularizers use the structure of the data itself. For example, in~\citep{sun2014structure}, the authors use the ``tag structure'' inherent in language data as regularization. Similarly, in~\citep{argyriou2008spectral}, the authors exploit multi-task structure in data for regularization. All of these methods either implicitly or explicitly relate regularization strength to the structure of the data. However, the additional computations during each training iteration for each of these methods are relatively demanding. In contrast, the approach we propose does not require any modifications to the underlying optimizer. The computational overhead comes from a one-time computation of the label softening strength before the model is trained.

\textbf{Bayes error rate (BER) estimation} Intuitively, regions of the feature space with high overlap may involve more complex decision boundaries as illustrated in Figure~\ref{Fig1:highlevel}(a) and (c); thus they have greater risk of overfitting. In this paper, we use a non-parametric bound on the BER to characterize the overlap between distributions and to derive our smoothing strength. Two recent papers~\citep{sekeh2018learning, berisha2016empirically} have shown that the BER can be bounded by Henze-Penrose (HP) divergence~\citep{henze1999multivariate}, which can be estimated by counting the number of edges connecting different classes in a Euclidean minimum spanning tree (MST). We use this divergence measure to devise the proposed structural regularizer adaptive to data. 

\textbf{Label smoothing, regularizer, or label noise?}~Label smoothing, a form of output distribution regularization prevents overfitting of a neural network by softening the ground-truth labels to penalize overconfident outputs~\citep{szegedy2016rethinking, pereyra2017regularizing}. The recent research~\citep{muller2019does} on label smoothing empirically reveals that it improves model calibration. As we show, label smoothing is similar to label flipping~\citep{xie2016disturblabel}; several papers have studied the negative effects of label flipping on model performance~\citep{natarajan2013learning, frenay2013classification, li2017noise, xiao2015learning}. In this paper, we quantify the distortion induced by the label smoothing and show that this bias is dependent on the statistical structure of the data. Then we devise an algorithm for smoothing such that this bias is mitigated.
\vspace{-0.2cm}
\section{Methodology}
Label smoothing inevitably distorts the training data distribution to penalize overconfident outputs. In this section, we first describe the uniform label smoothing (ULS) proposed in~\citep{szegedy2016rethinking} and study the distortion effects of the label smoothing by analyzing the bias this method induces with respect to the BER. Finally, using these insights, we devise a simple objective function to determine a distribution of smoothing strengths over the feature space such that the bias in the BER is minimized. 
\subsection{Uniform Label Smoothing (ULS)}
\label{subsec:3.1}
Given $N$ training examples $\mathbf{S}=\{\mathbf{x}_1,...,\mathbf{x}_N\}$ corresponding to a $K$-class classification problem, a neural network, parameterized by $\theta$, computes the probability $P_\theta(k|\mathbf{x})$ of label $k \in\{1,\ldots,K\}$ being assigned to training example $\mathbf{x}$ and compares this to the labels from the training data using a cross-entropy loss, 
\vspace{-0.35cm}
\begin{align}
    L(\theta)=-\frac{1}{N}\sum_{n=1}^N\sum_{k=1}^Kq(k|\mathbf{x}_n)\log(P_\theta(k|\mathbf{x}_n)),
    \label{eq:ULSloss}
\end{align}
where $q(k|\mathbf{x})$ is a one-hot encoding scheme for the labels and defined as $q(k|\mathbf{x})=\chi_{k,t}$; $\chi_{k,t}$ for ground-truth class label $t$ is an indicator that is $1$ if $k=t$ and $0$ otherwise. It has been shown that the one-hot distribution over the $K$ labels has the effect of training the network to be overconfident~\citep{szegedy2016rethinking,xie2016disturblabel,pereyra2017regularizing}. The authors in~\citep{szegedy2016rethinking} propose to smooth ground-truth labels by modifying the one-hot distribution to the smoothed label distribution $\hat{q}(k|\textbf{x})= (1-\epsilon)\chi_{k,t} + \epsilon u(k)$, where $\epsilon\in (0,1)$ is the smoothing strength and $u(k)=\frac{1}{K}$ is a fixed uniform distribution for all labels. 

We observe that this approach is similar to the label flipping regularizer~\citep{xie2016disturblabel} which modifies the ground-truth labels to one of the other classes chosen uniformly with probability $\epsilon\frac{1}{K}$. Said another way, this approach maintains the correct label with probability $(1-\alpha)$ and switches to one of the other labels with probability $\frac{\alpha}{K-1}$, where $\alpha = \epsilon \frac{K-1}{K}$. In this case, the expected distribution over labels for label flipping is exactly the same as in the label smoothing case.

\subsection{Bayes Error Rate (BER) Bias}
\label{subsec:3.2}
Given $K$ original class-conditional distributions in the training data $\{P(\mathbf{x}|y=1), P(\mathbf{x}|y=2),\ldots,P(\mathbf{x}|y=K)\}$ each with prior probability $P_k = P(y = k)$ and defining $[K]=\{1,\ldots,K\}$, the BER, $\mathcal{R}$, of the original data distributions is
\vspace{-0.35cm}
\begin{align}
    \mathcal{R} &= 1 - \int \max_{k\in[K]}P(y=k|\mathbf{x})P(\mathbf{x})d\mathbf{x},
    \label{eq:Bayes}
\end{align}
where $P(y=k|\mathbf{x}) = \frac{P_kP(\mathbf{x}|y=k)}{P(\mathbf{x})}$ is the posterior probability at $\mathbf{x}$ and $P(\mathbf{x})$ is the marginal distribution. For our analysis, it is helpful to define the point-wise BER,
\vspace{-0.35cm}
\begin{align}
    \mathcal{R}(\mathbf{x}) &= 1 - \max_{k\in[K]}P(y=k|\mathbf{x}).
    \label{eq:Pointbayes}
\end{align}
As discussed in Section~\ref{subsec:3.1}, label smoothing with uniform smoothing strength $\alpha$ is the same as flipping labels with probability $\alpha$ in expectation. We use this equivalence to derive our results. For the sake of generality, in our analysis we make the label flipping probability data dependent. That is, we define the probability of flipping a label for data point $\mathbf{x}$ as $\hat{\alpha}(\mathbf{x})$. Based on the equivalence of label smoothing and label flipping, the conditional posterior probability $\hat{P}(y=k|\mathbf{x})$ after label smoothing with data-dependent strength $\hat{\alpha}(\mathbf{x})$ is,
\vspace{-0.15cm}
\begin{align}
        \hat{P}(y=k|\mathbf{x})&= [1 - \hat{\alpha}(\mathbf{x})]P(y=k|\mathbf{x})\nonumber\\ 
        &+\hat{\alpha}(\mathbf{x})\frac{\sum_{i\neq k}P(y=i|\mathbf{x})}{K-1}
        \label{eq:smposterior}
\end{align}
This is simply a mixture of posteriors under the assumption that the label for training sample $\mathbf{x}$ is correct with probability $1-\hat{\alpha}(\mathbf{x})$ or flipped to any of the other classes with probability $\hat{\alpha}(\mathbf{x})$. If a label is flipped, each of the remaining classes are equally likely.

The smoothing factor, $\hat{\alpha}(\mathbf{x})$, controls the optimal decision boundary for the smoothed distribution. As we outlined in Section~\ref{sec:relatedwork}, there is a line of work~\citep{natarajan2013learning, frenay2013classification, li2017noise, xiao2015learning} that highlights the negative impact of aggressive label modification. We first begin by addressing the following natural question: If we let $T=\argmaxA_i\{P(y=i|\mathbf{x}))\}$, under what constraints on $\hat{\alpha}(\mathbf{x})$  can we ensure that $T=\argmaxA_i\{\hat{P}(y=i|\mathbf{x}))\}$? For this analysis, we analyze the difference between the optimal class  and all other classes for the smoothed posterior,
\vspace{-0.15cm}
\begin{align}
    &\hspace{0.4cm}\hat{P}(y=T|\mathbf{x})-\hat{P}_{k\neq T}(y=k|\mathbf{x})\nonumber\\
    &=[1-\hat{\alpha}(\mathbf{x})][P(y=T|\mathbf{x}) - P_{k\neq T}(y=k|\mathbf{x})]\nonumber\\
    &+\hat{\alpha}(\mathbf{x})\left[\frac{\sum_{i\neq T}P(y=i|\mathbf{x}) - \sum_{i\neq k}P(y=i|\mathbf{x})}{K-1}\right]\nonumber\\
    &=\left[1-\hat{\alpha}(\mathbf{x}) - \frac{\hat{\alpha}(\mathbf{x})}{K-1}\right]\underbrace{\left[P(y=T|\mathbf{x}) - P_{k\neq T}(y=k|\mathbf{x})\right]}_{>0}\nonumber\\
    &>0\Longrightarrow{}\hat{\alpha}(\mathbf{x}) < \frac{K-1}{K}
    \label{eq:posteriordiff}
\end{align}
This shows that having $\hat{\alpha}(\mathbf{x}) < \frac{K-1}{K}$ preserves the decision boundary after label smoothing since $\argmaxA_i\{\hat{P}(y=i|\mathbf{x}))\} = \argmaxA_i\{P(y=i|\mathbf{x}))\}$. This result makes intuitive sense since $\frac{K-1}{K}$ is also the threshold probability for the label flipping scheme to guarantee that the ``signal'' of the ground-truth labels is stronger than that of the noisy labels; e.g., for the binary case, the flipping probability should be less than $0.5$ to ensure more correct labels than incorrect labels for a class, thus preserving the correct optimal decision boundary. 

This is a rather loose constraint on the smoothing strength and it provides no information on how it should vary in different regions of the feature space. While this constraint preserves the decision boundary, it says nothing about how the smoothing distorts the amount of overlap between class distributions at any given point, $\mathbf{x}$. For insight into this, we analyze the BER. The BER after label smoothing, $\hat{\mathcal{R}}$, is
\vspace{-0.2cm}
\begin{align}
\hat{\mathcal{R}}(\mathbf{x}) &= 1 - \max_{i\in[k]}\hat{P}(y=i|\mathbf{x})
\label{eq:Pointbayessm}
\end{align}
The bias between the modified BER $\hat{\mathcal{R}}$ and the original BER $\mathcal{R}$ can then be derived from (\ref{eq:Pointbayes}), (\ref{eq:smposterior}), and (\ref{eq:Pointbayessm}) as follows, 
\vspace{-0.2cm}
\begin{align}
    &\hspace{0.4cm}\left|\mathcal{R} - \hat{\mathcal{R}}\right|\nonumber\\&= \int\left|\mathcal{R}(\mathbf{x}) - \hat{\mathcal{R}}(\mathbf{x})\right|P(\mathbf{x})d\mathbf{x}\nonumber\\&=\int\left|(1 - \hat{\alpha}(\mathbf{x}))P(y=T|\mathbf{x}) +\hat{\alpha}(\mathbf{x})\frac{\sum_{i\neq k}P(y=i|\mathbf{x})}{K-1}\right.\nonumber\\&\qquad-\left. P(y=T|\mathbf{x})\vphantom{\frac{1}{1}}\right|P(\mathbf{x})d\mathbf{x}\nonumber\\&=\int\hat{\alpha}(\mathbf{x})\left|\frac{1-P(y=T|\mathbf{x})}{K-1}-P(y=T|\mathbf{x})\right|P(\mathbf{x})d\mathbf{x}\nonumber\\
    &=\int \hat{\alpha}(\mathbf{x})\left|\mathcal{R}(\mathbf{x})\frac{K}{K - 1} - 1\right|P(\mathbf{x})d\mathbf{x}. \label{eq:risk-expression}
\end{align}
For the second equality, we use the fact that $T=\argmaxA_i\{P(y=i|\mathbf{x}))\} =\argmaxA_i\{\hat{P}(y=i|\mathbf{x}))\}$ if $\hat{\alpha}(\mathbf{x}) < \frac{K-1}{K}$. Then we integrate over the difference between the maximum posterior distributions - the original and the one after label smoothing.
From~\eqref{eq:risk-expression}, it is clear that if one retains  the original labels (i.e., smoothing with  $\hat{\alpha}(\mathbf{x})=0$), the BER is zero. On the other hand, if one uses uniform strength smoothing of  $\hat{\alpha}(\mathbf{x})=\alpha$ for all data points, then this imposes high bias at points with low original BER $\mathcal{R}(\mathbf{x})$ and high marginal likelihood $P(\mathbf{x})$. \vspace{-0.5cm}
\subsection{Structural Label Smoothing (SLS)}
\vspace{-0.15cm}
\label{subsec:3.3}
\vspace{-0.15cm}
\subsubsection{Choice of Smoothing Strength}
\vspace{-0.15cm}
\label{subsubsec:3.3.1}
In light of (\ref{eq:risk-expression}), we devise a simple objective function that will allow us to choose the smoothing strength $\hat{\alpha}(\mathbf{x})$ on labels for regularization as well as reduce the BER bias to realize structural label smoothing (SLS):
\vspace{-0.15cm}
\begin{align}
&\underset{\hat{\alpha}(\mathbf{x})}{\text{minimize}}
\int(\hat{\alpha}(\mathbf{x}) - \alpha)^2P(\mathbf{x})d\mathbf{x}\nonumber\\&+ \beta\int \hat{\alpha}(\mathbf{x})\left|\mathcal{R}(\mathbf{x})\frac{K}{K - 1} - 1\right|P(\mathbf{x})d\mathbf{x}\nonumber\\
&\text{subject to}
\int\hat{\alpha}(\mathbf{x})P(\mathbf{x})=\alpha
\label{eq:strengthcost}
\end{align}
where the first term and second term in the R.H.S. are the variance of the smoothing strength and the BER bias respectively. The user-defined parameters $\alpha$ and $\beta$ control the average smoothing strength and BER bias reduction strength. Indeed, computing the conditional BER $\mathcal{R}(\mathbf{x})$ for each data point is impossible for real-world data. We instead propose to estimate the BER at a cluster level using the algorithm described in Section~\ref{subsubsec:3.3.2} and replace the conditional BER $\mathcal{R}(\mathbf{x})$ at each data point with the BER $\mathcal{R}_c$ of the $c^{\text{th}}$ cluster to which the data point belongs. To this end, we rewrite the objective function in cluster form:
\vspace{-0.35cm}
\begin{align}
& \underset{\hat{\alpha}_c}{\text{minimize}}
\sum_c(\hat{\alpha}_c - \alpha)^2w_c + \beta\sum_c \hat{\alpha}_c\left|\mathcal{R}_c\frac{K}{K - 1} - 1\right|w_c\nonumber\\
& \text{subject to}
\sum_c\hat{\alpha}_cw_c=\alpha
\label{eq:strengthcostcluster}
\end{align}

\vspace{-0.5cm}
where $\hat{\alpha}_c$ stands for the smoothing strength imposed on the $c^{\text{th}}$ cluster, and $w_c$ estimates $P(\mathbf{x})$ using the ratio of the number of samples that fall in that cluster over all samples in the data. We can view clusters as cohesive regions of the feature space; our algorithm imposes a uniform smoothing strength within each cluster based on this cost function. 
The cost function has two terms and a constraint. The constraint imposes an average smoothing level, $\alpha$, across all samples. This first term in the cost function constrains the weighted variance of the smoothing strength such that it doesn't vary greatly from cluster to cluster. The second term in the cost function imposes the additional constraint to reduce the bias in the BER in each cluster. The cost function in \eqref{eq:strengthcostcluster} has a closed form solution given by 
\vspace{-0.3cm}
\begin{equation}
\hat{\alpha}_i\! = \!\alpha + \frac{\beta}{2} \left (\sum_c\left|R_c\frac{K}{K-1}-1\right|w_c- \left|R_i\frac{K}{K-1}-1\right|) \right).
\label{eq:closedformsolution}
\end{equation}
Observe that setting $\beta$ to zero results in $\hat{\alpha}_c = \alpha$, which is the traditional uniform label smoothing scheme. Increasing $\beta$ results in cluster-dependent smoothing that aims to reduce the BER subject to the average smoothing strength $\alpha$. All other things equal, the cost function above favors large $\hat{\alpha}_c$ for clusters with higher $\mathcal{R}_c$. This is aligned with the intuition that clusters with high $\mathcal{R}_c$ near the decision boundary require stronger regularization.
\vspace{-0.2cm}
\subsubsection{Estimation of Bayes Error Rate (BER)}
\label{subsubsec:3.3.2}
As is clear from Section~\ref{subsubsec:3.3.1}, SLS requires knowledge of the BER in different regions of the feature space. This is usually difficult to estimate. However, we will leverage a recent line of work~\citep{berisha2016empirically, sekeh2018learning} that proposes to bound the BER non-parametrically based on the Henze-Penrose (HP) divergence~\citep{henze1999multivariate}. The Henze-Penrose divergence $D_{ij}$ between class $i$ and $j$ with prior label probability $P_i$ and $P_j$ is given by  
\vspace{-0.15cm}
 \begin{align}
      D_{ij}&=\frac{1}{4P_{ij}P_{ji}}\left(\int\frac{P_{ij}P(\mathbf{x}|y=i)-P_{ji}P(\mathbf{x}|y=j)}{P_{ij}P(\mathbf{x}|y=i)+P_{ji}P(\mathbf{x}|y=j)}d\mathbf{x}\right.\nonumber\\&\left.-(P_{ij}-P_{ji})^2\vphantom{\frac{1}{1}}\right),
      \label{eq:HPdivergence}
 \end{align}
 where $P_{ij}=\frac{P_i}{P_i+P_j}$. There is a convenient upper and lower bound on the pair-wise BER provided by~\citep{berisha2016empirically} as follows    
 \vspace{-0.15cm}
 \begin{align}
    \frac{1}{2}-\frac{1}{2}\sqrt{u_{ij}} \leq\mathcal{R}_{ij}\leq\frac{1}{2}-\frac{1}{2}u_{ij}
    \label{eq:binarybound}
 \end{align}
 where $u_{ij}=4P_{ij}P_{ji}D_{ij}+(P_{ij} - P_{ji})^2$. For the binary case, where we denote the number of data points for two classes $i$ and $j$ by $N_i$ and $N_j$, the HP divergence can be estimated as follows. First, we construct a Euclidean minimal spanning tree (MST) on the data and count the number of edges that connect points from class $i$ to class $j$ (denoted as $C_{ij}$). It was shown in~\citep{berisha2016empirically} that  $1-C_{ij}\frac{N_i+N_j}{2N_iN_j}$ converges to $D_{ij}$. The authors in~\citep{sekeh2018learning} proposed a generalized bound for the multi-class case using only one global MST without having to perform pair-wise estimates of the BER,
 \vspace{-0.35cm}
 \begin{align}
     &\mathcal{R}\geq\frac{K-1}{K}\left[1-\left(1-2\frac{K}{K-1}\sum_{i=1}^{K-1}\sum_{j=i+1}^K\delta_{ij}\right)\right]^{1/2},\nonumber\\
     &\mathcal{R}\leq2\sum_{i=1}^{K-1}\sum_{j=i+1}^K\delta_{ij},
     \label{eq:multiclassbound}
 \end{align}
where $\delta_{ij}:=\int\frac{P_iP_jP(\mathbf{x}|y=i)P(\mathbf{x}|y=j)}{P(\mathbf{x})}d\mathbf{x}$. It was shown in~\citep{sekeh2018learning} that $\frac{C_{ij}}{N}\shortrightarrow\delta_{ij}$, where  $C_{ij}$ is the number of edges connecting class $i$ and $j$ in a global MST constructed on the size $N$ dataset. 

This approach provides a means of estimating a bound on the BER that we can use to derive the optimal smoothing strength for each cluster. There are two potential issues with the direct application of the bound above to our problem. First, the analysis above requires the use of all the data, whereas in our setting, it would suffice to estimate this quantity for different clusters in the data. Second, non-parametric methods have a glacially slow convergence rate for high-dimensional data~\citep{moon2015meta}. 

To address the first issue, we first cluster the data and compute the HP divergence for each cluster. The dataset is clustered into $C$ clusters $\mathbf{X}_c=\{\mathbf{X}_1,\ldots,\mathbf{X}_C\}$. Next, we estimate the BER bound of each cluster using either \eqref{eq:binarybound} for the binary case or \eqref{eq:multiclassbound} for the multi-class case. Lastly, we compute the label smoothing strength $\hat{\alpha}_c$ for each cluster using (\ref{eq:closedformsolution}) by replacing the BER with the estimated lower bound.
Now each cluster has a different smoothing strength derived such that the smoothing bias in the BER of each cluster is minimized.

The second issue implies that our estimate of the BER bound based on the HP divergence is potentially biased and imprecise for high-dimensional data. However, we empirically observe that the loss function in \eqref{eq:strengthcostcluster} is not very sensitive to this as long as the bounds above capture the trends by cluster (e.g. clusters with high overlap between class distributions have higher estimated BER bounds and vice versa). As we will see in Section~\ref{subsec:4.3}, even imprecise estimates of the BER for clusters are sufficient to mitigate the potentially negative consequences associated with label smoothing while maintaining its property as a regularizer. We believe a that a careful investigation of the efficacy of the proposed method is an interesting avenue for future work.
\vspace{-0.2cm}
\section{Experimental Results}
\vspace{-0.2cm}
In the following sections, we compare the performance of SLS to ULS with a series of experiments on synthetic data, Higgs, SVHN, CIFAR-10 and CIFAR-100 datasets. With the cluster-dependent smoothing strength $\hat{\alpha}_c$ for data cluster $\mathbf{X}_c$, we construct the training loss for SLS 
\begin{align}
    \tilde{L}(\theta)&=-\frac{1}{N}\sum_{c=1}^C\sum_{n_c=1}^{N_c}\sum_{k_{n_c=1}}^K\hat{q}_c(k_{n_c}|\mathbf{x}_{n_c})log(P_\theta(k_{n_c}|\textbf{x}_{n_c})),\nonumber\\&\hspace{4.5cm}\forall\mathbf{x}_{n_c}\in \mathbf{X}_c,
    \label{eq:losssm}
\end{align}
where $\hat{q}_c(k_{n_c}|\mathbf{x}_{n_c})=\begin{cases}\frac{\hat{\alpha}_c}{K-1}, k_{n_c}\neq t\\1-\hat{\alpha}_c, k_{n_c}=t \end{cases}$.
For each experiment, we set the average smoothing strength $\alpha$ such that it is the same for both SLS and ULS for a fair comparison.  
\vspace{-0.3cm}
\subsection{Experiments on Synthetic Data}
\vspace{-0.2cm}
For the synthetic dataset, we create a binary classification problem with two classes of data $S_0$ and $S_1$ generated from a two-dimensional Gaussian mixture model (GMM) $S_0\sim\sum_{i=1}^6\frac{1}{6}\mathcal{N}(\mathbf{u}_i^0, \mathbf{\Sigma}_i^0)$ and $S_1\sim\sum_{i=1}^6\frac{1}{6}\mathcal{N}(\mathbf{u}_i^1, \mathbf{\Sigma}_i^1)$ where $\mathbf{u}_i^0$ and $\mathbf{u}_i^1$ are mean vectors and $\mathbf{\Sigma}_i^0$ and $\mathbf{\Sigma}_i^1$ are covariance matrices of each component for class 0 and class 1. Subsequently, we add 126 noisy features with each sampled from a Gaussian model $\mathcal{N}(u, \sigma)$. In other words, we generate a synthetic 128-dimensional classification problem, where only 2 of the dimensions are useful for classification. We independently draw 12000 samples from this distribution for the training set and another 12000 for the test set. A full description of the parameters used to generate synthetic dataset is provided in the supplementary material. This dataset is visualized on the 2 useful dimensions in Figure~\ref{Fig1:highlevel}(a). 
\begin{figure}[h]
  \centering
  \includegraphics[width=0.8\linewidth]{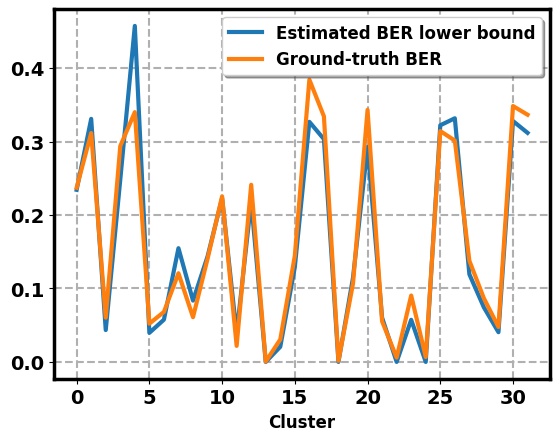}
 \vspace{-0.3cm}
\caption{\footnotesize{BER lower bound and the Ground-truth BER.}}
\label{Fig2:comp}
\end{figure}
To compute the cluster smoothing strength $\hat{\alpha}_c$ in \eqref{eq:strengthcostcluster}, we use principal component analysis (PCA) + GMM-based clustering to divide the training data into 32 clusters. For each cluster, we estimate the BER lower bound using \eqref{eq:binarybound}. Lastly, we compute the smoothing strength $\alpha_c$ for each cluster by replacing the BER in \eqref{eq:strengthcostcluster} with the estimated lower bound ranging the average smoothing strength $\alpha$ from $0.1$-$0.3$ and the strength of the BER reduction term $\beta$ from $0.1$-$0.5$. Both the ground-truth and estimated BER lower bound of each cluster are quantitatively shown in Figure~\ref{Fig2:comp}. As expected, the ``lower bound" does not always lower-bound the BER due to the finite-sample bias; however, it closely tracks the trend cluster by cluster.

After smoothing the labels with strength $\hat{\alpha}_c$, we train a single layer neural network with 256 hidden units for 2500 epochs with a learning rate of $0.01$ using the stochastic gradient descent (SGD) optimizer. We compare the results for models regularized by SLS to two baselines: training without any regularization and with the ULS.
\begin{table}[h]
\centering
\resizebox{0.35\textwidth}{!}{%
\renewcommand\tabcolsep{4pt} 
\begin{tabular}{|c|c|c|c|c|}
\hline
Regularization &$\alpha$ &$\beta$ &Mean$\pm$ std&Lowest\\ 
\hline
{\tt None} &-& -&24.82\rpm0.19&24.65
\\\hline
{\tt ULS} &0.1 &-&24.63\rpm0.14&24.50
\\
{\tt SLS} &0.1  &0.1&\textbf{24.60\rpm0.04}&24.56
\\
{\tt SLS} &0.1  &0.2&\textbf{24.55\rpm 0.22}&\textbf{24.20}
\\
{\tt SLS} &0.1  &0.3&\textbf{24.62\rpm0.15}&\textbf{24.40}
\\
{\tt SLS} &0.1  &0.4&\textbf{\textbf{24.48\rpm0.11}}&\textbf{24.29}\\
{\tt SLS} &0.1  &0.5&\textbf{24.52\rpm0.12}&\textbf{24.42}\\
{\tt RevSLS} &0.1  &0.1&24.92\rpm0.24&24.71\\
{\tt RevSLS} &0.1  &0.2&25.22\rpm0.23&24.89\\
{\tt RevSLS} &0.1  &0.3&25.08\rpm0.18&24.83\\ 
{\tt RevSLS} &0.1  &0.4&25.35\rpm0.19&25.17\\
{\tt RevSLS} &0.1  &0.5&25.73\rpm0.08&25.64\\\hline
{\tt ULS} &0.2&-&25.02\rpm0.29&24.59\\
{\tt SLS} &0.2  &0.1&\textbf{24.68\rpm0.11}&\textbf{24.58}\\
{\tt SLS} &0.2  &0.2&\textbf{24.42\rpm0.07}&\textbf{24.33}\\
{\tt SLS} &0.2  &0.3&\textbf{24.46\rpm0.05}&\textbf{24.38}\\
{\tt SLS} &0.2  &0.4&\textbf{24.45\rpm0.08}&\textbf{24.34}\\
{\tt SLS} &0.2  &0.5&\textbf{24.46\rpm0.11}&\textbf{24.28}\\ 
{\tt RevSLS} &0.2  &0.1&25.28\rpm0.25&24.91\\
{\tt RevSLS} &0.2  &0.2&25.26\rpm0.11&25.08\\ 
{\tt RevSLS} &0.2  &0.3& 25.43\rpm0.07&25.34\\
{\tt RevSLS} &0.2  &0.4& 25.47\rpm0.10&25.30\\
{\tt RevSLS} &0.2  &0.5& 25.62\rpm0.05&25.56\\\hline
{\tt ULS} &0.3 &-&25.67\rpm0.13&25.47\\
{\tt SLS} &0.3  &0.1&\textbf{24.97\rpm0.14}&\textbf{24.78}\\ 
{\tt SLS} &0.3  &0.2&\textbf{24.58\rpm0.08}&\textbf{24.48}\\
{\tt SLS} &0.3  &0.3&\textbf{24.55\rpm0.03}&\textbf{24.53}\\
{\tt SLS} &0.3  &0.4&\textbf{24.76\rpm0.09}&\textbf{24.65}\\
{\tt SLS} &0.3  &0.5&\textbf{24.75\rpm0.04}&\textbf{24.70}\\
{\tt RevSLS} &0.3  &0.1&25.61\rpm0.13&25.40\\ 
{\tt RevSLS} &0.3  &0.2&25.44\rpm0.17&25.27\\ 
{\tt RevSLS} &0.3  &0.3&25.63\rpm0.12&25.45\\ 
{\tt RevSLS} &0.3  &0.4&25.85\rpm0.05&25.85\\ 
{\tt RevSLS} &0.3  &0.5&26.03\rpm0.05&25.97\\
\hline
\end{tabular}
}
\caption{\footnotesize{Test error rate (\%) for the synthetic dataset with best algorithms highlighted in bold for each smoothing strength $\alpha$.}}
\vspace{-0.6cm}
\label{Tab1:Testsyn}
\end{table}
To further validate that the estimated smoothing strength in \eqref{eq:strengthcostcluster} is rational, we reverse the order of $\hat{\alpha}_c$ by swapping the large values of $\hat{\alpha}_c$ to the small values of $\hat{\alpha}_c$ and train the neural network with the reversed $\hat{\alpha}_c$ using exactly the same experimental settings as before. Our expectation is that this would negatively impact the results. The complete table of results is shown in Table~\ref{Tab1:Testsyn}, where SLS denotes structural label smoothing and RevSLS reversed structural label smoothing.

\begin{figure*}[h]
\centering
\vspace{-0.2cm}
\begin{subfigure}{.32\linewidth}
  \centering
  \includegraphics[width=1\linewidth]{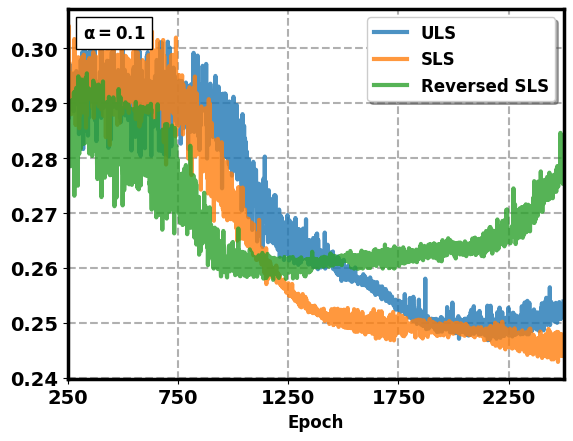}
  \vspace{-0.6cm}
\end{subfigure}
\begin{subfigure}{.32\linewidth}

  \centering
  \includegraphics[width=1\linewidth]{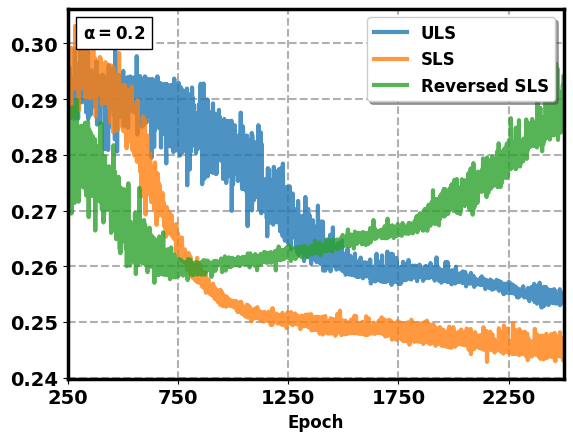}
  \vspace{-0.6cm}
\end{subfigure}
\begin{subfigure}{.32\linewidth}
  \centering
  \includegraphics[width=1\linewidth]{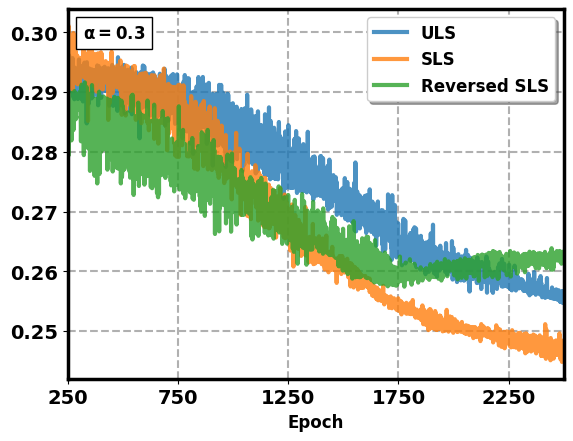}
  \vspace{-0.6cm}
\end{subfigure}
\vspace{-0.3cm}
\caption{\footnotesize{Test error curves for ULS, SLS and reversed SLS with various average/uniform smoothing strength $\alpha$.}}
\label{Fig3:Testcurvesyn}
\end{figure*}

From Table~\ref{Tab1:Testsyn} we observe that training with the SLS consistently outperforms both training with ULS and training without regularization for all hyperparameter values according to the average test error and the lowest test error. As further validation, training with reversed SLS increases the test error such that it is higher than both the corresponding training with ULS and training without regularization. To further show this, we plot the test error during training for ULS, SLS and reversed SLS in Figure~\ref{Fig3:Testcurvesyn} for three values of $\alpha$. The figure shows that SLS not only performs the best but also converges faster than ULS. As expected, the test error for the reversed SLS increases after a while as the original data distributions have been largely distorted.  

In section 5 of the supplementary material, we empirically evaluate the sensitivity of SLS on the number of clusters used to estimate the smoothing strengths. The results reveal that the approach is only mildly sensitive to the number of clusters; SLS consistently outperforms ULS for 16, 32, and 64 clusters.
\vspace{-0.4cm}
\subsection{Experiments on Real Data}
\vspace{-0.2cm}
\subsubsection{Binary classification}
\label{subsubsec:4.2.1}

\begin{figure}[h]
\centering
\vspace{-0.2cm}
\begin{subfigure}{.47\linewidth}
  \centering
  \includegraphics[width=1\linewidth]{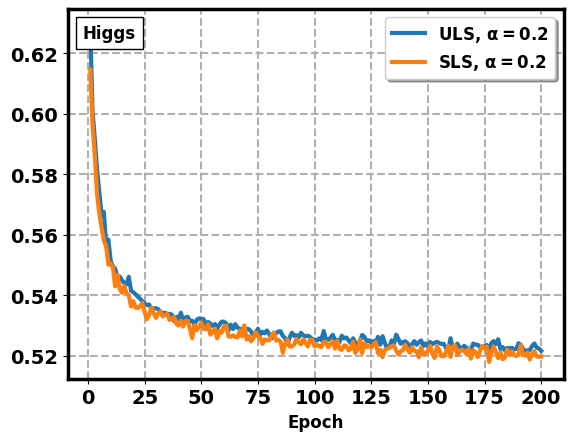}
\end{subfigure}
\begin{subfigure}{.47\linewidth}
  \centering
  \includegraphics[width=1\linewidth]{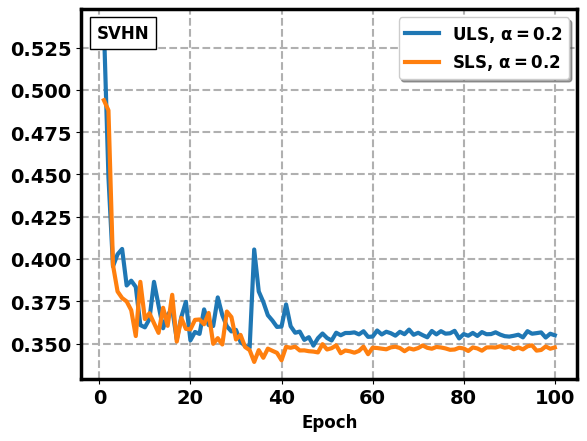}
\end{subfigure}\\
\begin{subfigure}{.47\linewidth}
  \centering
  \includegraphics[width=1\linewidth]{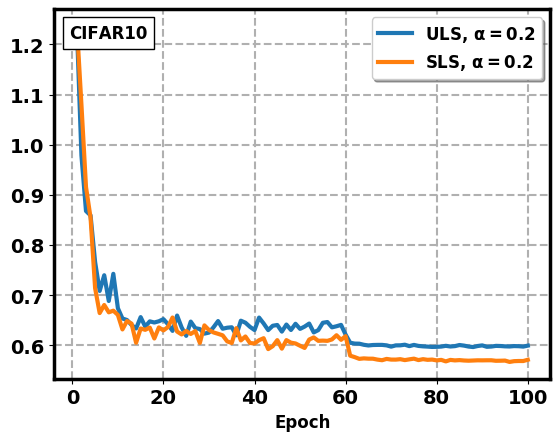}
\end{subfigure}
\begin{subfigure}{.47\linewidth}
  \centering
  \includegraphics[width=1\linewidth]{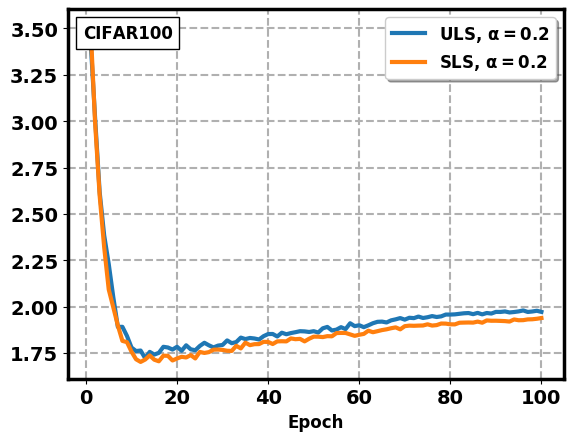}
\end{subfigure}
\vspace{-0.2cm}
\caption{\footnotesize{Cross-entropy on real test set along training (with batch-normalization) for ULS and SLS with $\alpha = 0.2$.}}
\label{Fig4:TestCE}
\vspace{-0.6cm}
\end{figure}
\textbf{Higgs}~\citep{baldi2014searching} contains 11M instances in two classes with 28 input features for a binary classification task. Following~\citep{baldi2014searching}, we use the last 500,000 instances of the dataset as a test set and the rest as a training set. We uniformly sample 1M instances from the training set, divide the samples to 128 clusters and use \eqref{eq:binarybound}  to estimate the BER lower bound. We use the estimated bound to compute the smoothing strength  $\hat{\alpha}_c$ for clusters of the sampled instances. As before, we use the same $\hat{\alpha}_c$ for all samples falling in the same cluster. We train a 7 layer CNN for 200 epochs with the same initialization for ULS and SLS and set the smoothing strength $\alpha=0.2$ which yielded the lowest error rate for both ULS and SLS. The cross-entropy of the test set for each training epoch for the Higgs data is shown in Figure~\ref{Fig4:TestCE}. The cross-entropy for SLS is consistently lower and results in a \textbf{0.2\%} error decrement over the ULS from 23.0\% to 22.8\%. A thorough examination of the SLS is reported in section~\ref{subsubsec:4.2.2} for a more complex multiclass classification task.
\vspace{-0.2cm}
\subsubsection{Multiclass classification}
\label{subsubsec:4.2.2}
\vspace{-0.2cm}
\textbf{SVHN}~\citep{netzer2011reading} consists of 73257 training digit images and 26032 testing digit images in 10 classes with a size of $32\times32$.  \textbf{CIFAR-10} and \textbf{CIFAR-100}~\citep{krizhevsky2009learning} consist of 60000 $32\times32$ images in 10 classes and 100 classes respectively with 50000 images for training and 10000 images for testing. 
\begin{table*}[h]
\renewcommand\tabcolsep{4pt} 
\centering
\begin{subtable}{0.65\textwidth}
\resizebox{1\textwidth}{!}{
\begin{tabular}{|ccc|cc|}
\hline
\multirow{2}{*}{\footnotesize{Regularization}}&\multirow{2}{*}{$\alpha$}&\multirow{2}{*}{$\beta$}& \multicolumn{2}{c|}{SVHN}\\\cline{4-5}
& &&Mean$\pm$ std&Lowest\\
\hline
{\tt None} & -&-& 5.49\rpm0.18&5.27\\\hline
{\tt ULS} &0.1&-&5.21\rpm 0.10&5.11\\
{\tt SLS} &0.1&0.2&\textbf{5.11\rpm 0.14}&\textbf{4.92}\\\hline
{\tt ULS} &0.2&-&5.30\rpm 0.04 &5.25\\
{\tt SLS} &0.2 &0.6&\textbf{5.19\rpm0.14} & \textbf{5.01}\\\hline
{\tt ULS} &0.3&-& 5.24\rpm0.09&5.10\\
{\tt SLS} &0.3&0.4&\textbf{5.15\rpm0.11}&\textbf{5.02}\\\hline
{\tt Batchnorm+} &&&&\\\hline
{\tt None} & -&-& 4.28\rpm 0.02 &4.26\\\hline
{\tt ULS} &0.1&-&4.01\rpm 0.09&3.95\\
{\tt SLS} &0.1&0.2&\textbf{3.97\rpm 0.04} &\textbf{3.91}\\\hline
{\tt ULS} &0.2&-&4.12\rpm 0.12 &3.95\\
{\tt SLS} &0.2&0.4&\textbf{4.01\rpm 0.11}&\textbf{3.91}\\\hline
{\tt ULS} &0.3&-& 4.01\rpm0.12&3.90\\
{\tt SLS} &0.3&0.4&\textbf{4.00\rpm 0.10}&\textbf{3.88}\\\hline
\end{tabular}
\begin{tabular}{|ccc|cc|}
\hline
\multirow{2}{*}{\footnotesize{Regularization}}&\multirow{2}{*}{$\alpha$}&\multirow{2}{*}{$\beta$}& \multicolumn{2}{c|}{CIFAR-10}\\\cline{4-5}
& &&Mean$\pm$ std&Lowest\\
\hline
{\tt None} & -&-& 22.43\rpm1.66 &20.62\\\hline
{\tt ULS} &0.1&-&21.07\rpm1.52&\textbf{19.12}\\
{\tt SLS} &0.1&0.4&\textbf{20.64\rpm1.11}&19.16\\\hline
{\tt ULS} &0.2&-&21.77\rpm1.64&19.89\\
{\tt SLS} &0.2&0.8&\textbf{21.22\rpm1.04}&\textbf{19.62}\\\hline
{\tt ULS} &0.3&-&21.81\rpm1.52&20.64\\
{\tt SLS} &0.3&0.6&\textbf{21.14\rpm0.56}&\textbf{20.35}\\\hline
{\tt Batchnorm+} &&&&\\\hline
{\tt None} & -&-&12.34\rpm0.27&11.99\\\hline
{\tt ULS} &0.1&-&\textbf{12.49\rpm0.25}&12.16\\
{\tt SLS} &0.1&0.8&12.52\rpm0.38 &\textbf{11.90}\\\hline
{\tt ULS} &0.2&-&12.57\rpm0.18&12.39\\
{\tt SLS} &0.2&0.2&\textbf{12.23\rpm0.32}&\textbf{11.90}\\\hline
{\tt ULS} &0.3&-&12.92\rpm0.22&12.71\\
{\tt SLS} &0.3&0.8&\textbf{12.74\rpm0.24}&\textbf{12.46}\\\hline
\end{tabular}
\begin{tabular}{|ccc|cc|}
\hline
\multirow{2}{*}{\footnotesize{Regularization}}&\multirow{2}{*}{$\alpha$}&\multirow{2}{*}{$\beta$}& \multicolumn{2}{c|}{CIFAR-100}\\\cline{4-5}
& &&Mean$\pm$ std&Lowest\\
\hline
{\tt None} & -&-& 57.39\rpm2.06 &55.48\\\hline
{\tt ULS} &0.1&-&58.89\rpm3.42&55.52\\
{\tt SLS} &0.1&0.6&\textbf{56.27\rpm1.55} &\textbf{55.16}\\\hline
{\tt ULS} &0.2&-&\textbf{55.09\rpm1.13}&\textbf{53.46}\\
{\tt SLS} &0.2&0.8&56.85\rpm2.14& 55.01\\\hline
{\tt ULS} &0.3&-&54.26\rpm1.44&53.23\\
{\tt SLS} &0.3&0.8&\textbf{54.00\rpm2.23}&\textbf{52.26}\\\hline
{\tt Batchnorm+} &&&&\\\hline
{\tt None} & -&-&37.83\rpm0.24&37.45\\\hline
{\tt ULS} &0.1&-&38.30\rpm0.82&37.26\\
{\tt SLS} &0.1&0.4&\textbf{38.03\rpm0.85} &\textbf{37.16}\\\hline
{\tt ULS} &0.2&-&37.77\rpm0.50&37.22\\
{\tt SLS} &0.2&0.4&\textbf{37.24\rpm0.70}&\textbf{36.64}\\\hline
{\tt ULS} &0.3&-&38.25\rpm0.31&37.82\\
{\tt SLS} &0.3&0.6&\textbf{38.13\rpm0.57}&\textbf{37.20}\\\hline
\end{tabular}
}
\caption{}
\label{Tab2a:testerror}
\end{subtable}
\quad
\begin{subtable}{0.26\textwidth}
\resizebox{1\textwidth}{!}{
\begin{tabular}{|cc|ccc|}
\hline
\footnotesize{Regularization}&$\alpha$&SVHN&CIFAR-10&CIFAR-100\\
{\tt None} & -&0.7905& 2.3823 &8.5204\\\hline
{\tt ULS} & 0.1&0.2922&0.7613&2.6632\\
{\tt SLS} & 0.1&\textbf{0.2904}&\textbf{0.7415}&\textbf{2.4696}\\\hline
{\tt ULS} & 0.2&0.3881&0.8394&\textbf{2.4891}\\
{\tt SLS} & 0.2&\textbf{0.3880}&\textbf{0.8144}&2.5210\\\hline
{\tt ULS} & 0.3&\textbf{0.5019}&0.9264&2.5606\\
{\tt SLS} & 0.3&0.5040&\textbf{0.9091}&\textbf{2.4959}\\\hline
{\tt Batchnorm+} &&&&\\\hline
{\tt None} & -&0.3353&0.8431&2.7568\\\hline
{\tt ULS} & 0.1&0.2411&0.5243&1.8344\\
{\tt SLS} & 0.1&\textbf{0.2407}&\textbf{0.5205}&\textbf{1.8062}\\\hline
{\tt ULS} & 0.2&0.3480&0.5900&1.8834\\
{\tt SLS} & 0.2&\textbf{0.3427}&\textbf{0.5822}&\textbf{1.8621}\\\hline
{\tt ULS} & 0.3&\textbf{0.4694}&0.6953&1.9945\\
{\tt SLS} & 0.3&0.4700&\textbf{0.6897}&\textbf{1.9790}\\\hline
\end{tabular}}
\caption{}
\label{Tab2b:crossentropy}
\end{subtable}
\vspace{-0.4cm}
\caption{\footnotesize{(a) Test error rates (\%) and (b) cross-entropy of training without regularization, with ULS and SLS for the real datasets. The best performing algorithms are highlighted in bold for each smoothing strength $\alpha$.}}
\label{Tab2:Testreal}
\vspace{-0.65cm}
\end{table*}
We use the MobileNet-v2~\citep{sandler2018mobilenetv2} as a backbone architecture and train on the three datasets 100 epochs without regularization, with ULS and with SLS for five times. We range $\alpha$ from $0.1$-$0.3$ for ULS and SLS. For the training without batch normalization, we use the Adam optimizer with a learning rate $0.001$ for the first 70 epochs and $0.0001$ for the rest; for the training with batch normalization, we use the SGD optimizer with a learning rate of $0.1$ for the first 70 epochs and $0.02$ for the rest combined with momentum of $0.9$.

To estimate the BER bounds, we first train a simple autoencoder for each dataset to reduce the data dimension to 256. Then we divide the training data for SVHN, CIFAR-10 and CIFAR-100 to 80, 64 and 48 clusters using $k$-Means clustering. Datasets with fewer samples per class are divided into a smaller number of clusters for better estimation of the BER lower bound. In the multiclass case, we use \eqref{eq:multiclassbound} for BER bounds estimation. Lastly, we compute $\hat{\alpha}_c$ with $\alpha$ ranging from $0.1$-$0.3$ and $\beta$ ranging from $0.2$-$1$. We train the network regularized by SLS using the same setting as the ULS training. For each value of $\alpha$, we train the network once for all values of $\beta$. For the $\beta$ with the lowest error rate in that run, we further train the network an additional four times. By evaluating on test data, we report the mean and standard deviation of accuracy in Table~\ref{Tab2a:testerror} and the cross-entropy on the test data in Table~\ref{Tab2b:crossentropy}. 
 
In Table~\ref{Tab2a:testerror}, we observe that training with SLS generally has a lower test error rate than the model without regularization and the model with ULS in terms of the average test error and lowest test error. As expected, the effect size is reduced after adding batch normalization.  With the same hyperparameters as in Table~\ref{Tab2a:testerror}, we also evaluated cross-entropy of the models on the test data. The results shown in Table~\ref{Tab2b:crossentropy} indicate that SLS achieves a lower cross-entropy and, as a result, is potentially more robust than the model with ULS and the model without regularization. Representative trends of the cross-entropy shown in Figure~\ref{Fig4:TestCE} reveal that the cross-entropy for the model with SLS is consistently lower than the model with ULS along the training. Complete results are shown in the Section 2 of the supplementary material.

To show that these improvements hold in other applications, we also compare SLS to ULS on a phoneme classification task using a subset of the TIMIT corpus. Both the test error rate and the analysis of cluster number are shown in the Section 6 of supplementary material. Consistent with other results, SLS consistently outperforms ULS.
\vspace{-0.4cm}
\subsection{Additional Evidence for the Rationale Behind SLS}
\label{subsec:4.3}
\vspace{-0.2cm}
We infer from (\ref{eq:strengthcostcluster}) that the label smoothing strength imposed on each cluster of the training data is supposed to positively correlate with the BER to reduce the BER bias. In this section, we provide additional evidence to further show this and to show that the classifier learns a more complex decision boundary in clusters where the BER is higher. 

The authors in~\citep{arpit2017closer} introduce the critical samples ratio (CSR) to describe the complexity of the decision boundary near samples. We estimate the CSR by searching the proximate space of a training example $\mathbf{x}$ and label it a \emph{critical sample} if there exists at least one adversarial example $\mathbf{\hat{x}}$ nearby. Formally, $\mathbf{x}$ is a critical sample if $\argminA_iP_\theta(y=i|\mathbf{x})\neq\argminA_jP_\theta(y=j|\mathbf{\hat{x}})$ subject to $||\mathbf{x}-\mathbf{\hat{x}}||_\infty<r$ where $r$ is the size of the box around $\mathbf{x}$ and $P_\theta(y|\mathbf{x})$ is the model prediction probability parameterized by $\theta$. The intuition behind the CSR is that the more critical samples exist in a dataset $\mathcal{D}$, the more complex the learned decision boundary by the classifier.

We compute the CSR for each cluster of the SVHN, CIFAR-10 and CIFAR-100 training sets by the Langevin adversarial sample search (LASS) algorithm in~\citep{arpit2017closer} to examine the complexity of the decision boundary learned by MobileNet-v2 without any regularization. Furthermore, for the clusterings of SVHN, CIFAR-10 and CIFAR-100 previously described, we compute the error difference (ED) between test and training error, the BER bound, and the SLS for three values of $\alpha$.

In Table~\ref{Tab4:Corr}, we show the pairwise correlation between each of these values. We observe a relatively strong positive correlation between the ED and the training data CSR. This indicates that a higher complexity decision boundary may be at higher risk for overfitting, thus requires more regularization. 

The estimated BER and the estimated smoothing strengths also both positively correlate with the CSR as shown in Table~\ref{Tab4:Corr}. This provides additional evidence that SLS applies stronger regularization to regions of the feature space where the learned decision boundary is more complex. This also provides evidence that our non-parametric estimator of the BER bound is useful, despite the fact that it likely has a large bias. The high correlation shows that it seems to capture the correct trend by cluster; it is this trend that is important for learning the optimal smoothing in each cluster.

\begin{table}[t]
\centering
\renewcommand\tabcolsep{4pt} 
\resizebox{0.4\textwidth}{!}{
\begin{tabular}{|c|c|c|c|c|c|c|}
\cline{2-7}
\multicolumn{1}{c|}{\textbf{SVHN}}&CSR&Test&BER&SLS1&SLS2&SLS3\\\hline
CSR&1&0.6091&0.7704&0.6785&0.7503&0.7301\\\hline
ED&0.6901&1&0.5329&0.5004&0.5350&0.5312\\\hline
BER&0.7704&0.5329&1&0.8964&0.9196&0.9168\\\hline
SLS1&0.6785&0.5004&0.8964&1&0.9687&0.9671\\\hline
SLS2&0.7503&0.5350&0.9196&0.9687&1&0.9949\\\hline
SLS3&0.7301&0.5312&0.9168&0.9671&0.9949&1\\\hline
\end{tabular}}\linebreak
\resizebox{0.4\textwidth}{!}{
\begin{tabular}{|c|c|c|c|c|c|c|}
\cline{2-7}
\multicolumn{1}{c|}{\textbf{CIFAR-10}}&CSR&Test&BER&SLS1&SLS2&SLS3\\\hline
CSR&1&0.7574&0.7062&0.6911&0.5857&0.6847\\\hline
ED&0.7574&1&0.5401&0.5300&0.4348&0.5388\\\hline
BER&0.7052&0.5401&1&0.9923&0.9100&0.9911\\\hline
SLS1&0.6911&0.5300&0.9923&1&0.9253&0.9976\\\hline
SLS2&0.5857&0.4348&0.9100&0.9253&1&0.9196\\\hline
SLS3&0.6847&0.5388&0.9911&0.9976&0.9196&1\\\hline
\end{tabular}}\linebreak
\resizebox{0.4\textwidth}{!}{
\begin{tabular}{|c|c|c|c|c|c|c|}
\cline{2-7}
\multicolumn{1}{c|}{\textbf{CIFAR-100}}&CSR&Test&BER&SLS1&SLS2&SLS3\\\hline
CSR&1&0.7980&0.8500&0.6958&0.5803&0.7034\\\hline
ED&0.7980&1&0.8127&0.7606&0.6975&0.7657\\\hline
BER&0.8500&0.8127&1&0.8722&0.7634&0.8905\\\hline
SLS1&0.6958&0.7606&0.8722&1&0.9591&0.9942\\\hline
SLS2&0.5803&0.6975&0.7634&0.9591&1&0.9425\\\hline
SLS3&0.7034&0.7657&0.8905&0.9942&0.9425&1\\\hline
\end{tabular}}
\caption{\footnotesize{Correlation matrices among CSR, error difference (ED) between test error and training error, estimated BER lower bound and selected smoothing strengths 0.1 (SLS1), 0.2 (SLS2) and 0.3 (SLS3) of clusters for the real datasets.}}
\label{Tab4:Corr}
\vspace{-0.65cm}
\end{table}
\vspace{-0.4cm}
\section{Conclusions}
\vspace{-0.3cm}
We propose a novel data-dependent regularization scheme called  Structural Label Smoothing (SLS). SLS is devised with the goal of using the data distribution to mitigate adverse effects of label modification, while retaining favorable regularization properties. 
Our regularization protocol clusters the training dataset, estimates the BER of each cluster, and learns  the smoothing strength for  each cluster. Experiments on both synthetic and real datasets show the effectiveness of our proposed SLS method.
\section*{Acknowledgement}
This research was supported in part by the Office of Naval Research grant N000141410722.
\bibliographystyle{plainnat}
\bibliography{egbib}

\begin{thebibliography}{29}
\providecommand{\natexlab}[1]{#1}
\providecommand{\url}[1]{\texttt{#1}}
\expandafter\ifx\csname urlstyle\endcsname\relax
  \providecommand{\doi}[1]{doi: #1}\else
  \providecommand{\doi}{doi: \begingroup \urlstyle{rm}\Url}\fi

\bibitem[Argyriou et~al.(2008)Argyriou, Pontil, Ying, and
  Micchelli]{argyriou2008spectral}
Andreas Argyriou, Massimiliano Pontil, Yiming Ying, and Charles~A Micchelli.
\newblock A spectral regularization framework for multi-task structure
  learning.
\newblock In \emph{Advances in Neural Information Processing Systems}, pages
  25--32, 2008.

\bibitem[Arpit et~al.(2017)Arpit, Jastrz{\k{e}}bski, Ballas, Krueger, Bengio,
  Kanwal, Maharaj, Fischer, Courville, Bengio, et~al.]{arpit2017closer}
Devansh Arpit, Stanis{\l}aw Jastrz{\k{e}}bski, Nicolas Ballas, David Krueger,
  Emmanuel Bengio, Maxinder~S Kanwal, Tegan Maharaj, Asja Fischer, Aaron
  Courville, Yoshua Bengio, et~al.
\newblock A closer look at memorization in deep networks.
\newblock \emph{arXiv preprint arXiv:1706.05394}, 2017.

\bibitem[Baldi et~al.(2014)Baldi, Sadowski, and Whiteson]{baldi2014searching}
Pierre Baldi, Peter Sadowski, and Daniel Whiteson.
\newblock Searching for exotic particles in high-energy physics with deep
  learning.
\newblock \emph{Nature communications}, 5:\penalty0 4308, 2014.

\bibitem[Berisha et~al.(2016)Berisha, Wisler, Hero~III, and
  Spanias]{berisha2016empirically}
Visar Berisha, Alan Wisler, Alfred~O Hero~III, and Andreas Spanias.
\newblock Empirically estimable classification bounds based on a nonparametric
  divergence measure.
\newblock \emph{IEEE Trans. Signal Processing}, 64\penalty0 (3):\penalty0
  580--591, 2016.

\bibitem[Fr{\'e}nay and Verleysen(2013)]{frenay2013classification}
Beno{\^\i}t Fr{\'e}nay and Michel Verleysen.
\newblock Classification in the presence of label noise: a survey.
\newblock \emph{IEEE transactions on neural networks and learning systems},
  25\penalty0 (5):\penalty0 845--869, 2013.

\bibitem[Henze et~al.(1999)Henze, Penrose, et~al.]{henze1999multivariate}
Norbert Henze, Mathew~D Penrose, et~al.
\newblock On the multivariate runs test.
\newblock \emph{The Annals of Statistics}, 27\penalty0 (1):\penalty0 290--298,
  1999.

\bibitem[Ioffe and Szegedy(2015)]{ioffe2015batch}
Sergey Ioffe and Christian Szegedy.
\newblock Batch normalization: Accelerating deep network training by reducing
  internal covariate shift.
\newblock \emph{arXiv preprint arXiv:1502.03167}, 2015.

\bibitem[Kannan et~al.(2018)Kannan, Kurakin, and
  Goodfellow]{kannan2018adversarial}
Harini Kannan, Alexey Kurakin, and Ian Goodfellow.
\newblock Adversarial logit pairing.
\newblock \emph{arXiv preprint arXiv:1803.06373}, 2018.

\bibitem[Krizhevsky and Hinton(2009)]{krizhevsky2009learning}
Alex Krizhevsky and Geoffrey Hinton.
\newblock Learning multiple layers of features from tiny images.
\newblock Technical report, Citeseer, 2009.

\bibitem[Krogh and Hertz(1992)]{krogh1992simple}
Anders Krogh and John~A Hertz.
\newblock A simple weight decay can improve generalization.
\newblock In \emph{Advances in Neural Information Processing Systems}, pages
  950--957, 1992.

\bibitem[Li et~al.(2017)Li, Qian, and Ji]{li2017noise}
Weizhi Li, Xiaoning Qian, and Jim Ji.
\newblock Noise-tolerant deep learning for histopathological image
  segmentation.
\newblock In \emph{2017 IEEE International Conference on Image Processing
  (ICIP)}, pages 3075--3079. IEEE, 2017.

\bibitem[Moon et~al.(2015)Moon, Hero, and Delouille]{moon2015meta}
Kevin~R Moon, Alfred~O Hero, and B~V{\'e}ronique Delouille.
\newblock Meta learning of bounds on the bayes classifier error.
\newblock In \emph{2015 IEEE Signal Processing and Signal Processing Education
  Workshop (SP/SPE)}, pages 13--18. IEEE, 2015.

\bibitem[Moosavi-Dezfooli et~al.(2018)Moosavi-Dezfooli, Fawzi, Uesato, and
  Frossard]{moosavi2018robustness}
Seyed-Mohsen Moosavi-Dezfooli, Alhussein Fawzi, Jonathan Uesato, and Pascal
  Frossard.
\newblock Robustness via curvature regularization, and vice versa.
\newblock \emph{arXiv preprint arXiv:1811.09716}, 2018.

\bibitem[M{\"u}ller et~al.(2019)M{\"u}ller, Kornblith, and
  Hinton]{muller2019does}
Rafael M{\"u}ller, Simon Kornblith, and Geoffrey Hinton.
\newblock When does label smoothing help?
\newblock \emph{arXiv preprint arXiv:1906.02629}, 2019.

\bibitem[Natarajan et~al.(2013)Natarajan, Dhillon, Ravikumar, and
  Tewari]{natarajan2013learning}
Nagarajan Natarajan, Inderjit~S Dhillon, Pradeep~K Ravikumar, and Ambuj Tewari.
\newblock Learning with noisy labels.
\newblock In \emph{Advances in Neural Information Processing Systems}, pages
  1196--1204, 2013.

\bibitem[Netzer et~al.(2011)Netzer, Wang, Coates, Bissacco, Wu, and
  Ng]{netzer2011reading}
Yuval Netzer, Tao Wang, Adam Coates, Alessandro Bissacco, Bo~Wu, and Andrew~Y
  Ng.
\newblock Reading digits in natural images with unsupervised feature learning.
\newblock 2011.

\bibitem[Pereyra et~al.(2017)Pereyra, Tucker, Chorowski, Kaiser, and
  Hinton]{pereyra2017regularizing}
Gabriel Pereyra, George Tucker, Jan Chorowski, {\L}ukasz Kaiser, and Geoffrey
  Hinton.
\newblock Regularizing neural networks by penalizing confident output
  distributions.
\newblock \emph{arXiv preprint arXiv:1701.06548}, 2017.

\bibitem[Sandler et~al.(2018)Sandler, Howard, Zhu, Zhmoginov, and
  Chen]{sandler2018mobilenetv2}
Mark Sandler, Andrew Howard, Menglong Zhu, Andrey Zhmoginov, and Liang-Chieh
  Chen.
\newblock Mobilenetv2: Inverted residuals and linear bottlenecks.
\newblock In \emph{Proceedings of the IEEE Conference on Computer Vision and
  Pattern Recognition}, pages 4510--4520, 2018.

\bibitem[Scardapane et~al.(2017)Scardapane, Comminiello, Hussain, and
  Uncini]{scardapane2017group}
Simone Scardapane, Danilo Comminiello, Amir Hussain, and Aurelio Uncini.
\newblock Group sparse regularization for deep neural networks.
\newblock \emph{Neurocomputing}, 241:\penalty0 81--89, 2017.

\bibitem[Sekeh et~al.(2018)Sekeh, Oselio, and Hero]{sekeh2018learning}
Salimeh~Yasaei Sekeh, Brandon Oselio, and Alfred~O Hero.
\newblock Learning to bound the multi-class bayes error.
\newblock \emph{arXiv preprint arXiv:1811.06419}, 2018.

\bibitem[Srivastava et~al.(2014)Srivastava, Hinton, Krizhevsky, Sutskever, and
  Salakhutdinov]{srivastava2014dropout}
Nitish Srivastava, Geoffrey Hinton, Alex Krizhevsky, Ilya Sutskever, and Ruslan
  Salakhutdinov.
\newblock Dropout: a simple way to prevent neural networks from overfitting.
\newblock \emph{The Journal of Machine Learning Research}, 15\penalty0
  (1):\penalty0 1929--1958, 2014.

\bibitem[Sun(2014)]{sun2014structure}
Xu~Sun.
\newblock Structure regularization for structured prediction.
\newblock In \emph{Advances in Neural Information Processing Systems}, pages
  2402--2410, 2014.

\bibitem[Szegedy et~al.(2016)Szegedy, Vanhoucke, Ioffe, Shlens, and
  Wojna]{szegedy2016rethinking}
Christian Szegedy, Vincent Vanhoucke, Sergey Ioffe, Jon Shlens, and Zbigniew
  Wojna.
\newblock Rethinking the inception architecture for computer vision.
\newblock In \emph{Proceedings of the IEEE conference on computer vision and
  pattern recognition}, pages 2818--2826, 2016.

\bibitem[Tibshirani(1996)]{tibshirani1996regression}
Robert Tibshirani.
\newblock Regression shrinkage and selection via the lasso.
\newblock \emph{Journal of the Royal Statistical Society: Series B
  (Methodological)}, 58\penalty0 (1):\penalty0 267--288, 1996.

\bibitem[Xiao et~al.(2015)Xiao, Xia, Yang, Huang, and Wang]{xiao2015learning}
Tong Xiao, Tian Xia, Yi~Yang, Chang Huang, and Xiaogang Wang.
\newblock Learning from massive noisy labeled data for image classification.
\newblock In \emph{Proceedings of the IEEE conference on computer vision and
  pattern recognition}, pages 2691--2699, 2015.

\bibitem[Xie et~al.(2016)Xie, Wang, Wei, Wang, and Tian]{xie2016disturblabel}
Lingxi Xie, Jingdong Wang, Zhen Wei, Meng Wang, and Qi~Tian.
\newblock Disturblabel: Regularizing cnn on the loss layer.
\newblock In \emph{Proceedings of the IEEE Conference on Computer Vision and
  Pattern Recognition}, pages 4753--4762, 2016.

\bibitem[Xue et~al.(2011)Xue, Chen, and Yang]{xue2011structural}
Hui Xue, Songcan Chen, and Qiang Yang.
\newblock Structural regularized support vector machine: a framework for
  structural large margin classifier.
\newblock \emph{IEEE Transactions on Neural Networks}, 22\penalty0
  (4):\penalty0 573--587, 2011.

\bibitem[Yuan and Lin(2006)]{yuan2006model}
Ming Yuan and Yi~Lin.
\newblock Model selection and estimation in regression with grouped variables.
\newblock \emph{Journal of the Royal Statistical Society: Series B (Statistical
  Methodology)}, 68\penalty0 (1):\penalty0 49--67, 2006.

\bibitem[Zhai and Wang(2018)]{zhai2018adaptive}
Ke~Zhai and Huan Wang.
\newblock Adaptive dropout with rademacher complexity regularization.
\newblock 2018.

\end{thebibliography}
\end{document}